# OBJECT SIEVING AND MORPHOLOGICAL CLOSING TO REDUCE FALSE DETECTIONS IN WIDE-AREA AERIAL IMAGERY

*Xin Gao, Sundaresh Ram, and Jeffrey J. Rodríguez*

Department of Electrical and Computer Engineering, The University of Arizona, Tucson, AZ, USA

**Abstract**

**For object detection in wide-area aerial imagery, post-processing is usually needed to reduce false detections. We propose a two-stage post-processing scheme which comprises an area-thresholding sieving process and a morphological closing operation. We use two wide-area aerial videos to compare the performance of five object detection algorithms in the absence and in the presence of our post-processing scheme. The automatic detection results are compared with the ground-truth objects. Several metrics are used for performance comparison.**

## 1. INTRODUCTION

In wide-area aerial imagery [1, 15], object detection methods [2, 3, 4, 5, 6] applied to low-resolution datasets often yield detection results with high percentage of wrong classification despite some post-processing that is done in those algorithms. Hence, a more effective post-processing technique is necessary to improve detection performance. Four existing post-processing schemes are briefly summarized as follows:

*1) Filtered dilation*: The scheme proposed by Salem *et al.* [11] uses a median filter for smoothing, and a dilation operator to shape the detected objects. It aims for distinct auto-detected regions with smooth borders and with small holes filled. It was originally specified as binary masking.

*2) Heuristic filtering*: Samarabandu and Liu [8] use two adjustable thresholds to discard unnecessary false detections after region extraction. One is an area threshold to exclude any detections smaller than 5% of the largest object in the ground truth; the other is an aspect threshold to exclude any detections having aspect ratio smaller than 0.2.

*3) Filtering by shape index*: The shape index (SI) introduced by Sharama *et al.* [12] is found by calculating the ratio of a region's perimeter to the square root of its area, then dividing this ratio by 4. After region segmentation, any detection with SI below the lowest SI of ground truth objects is excluded.

*4) Sieving large objects and removing small objects with morphological opening*: Zheng *et al.* [5] select an area threshold of 2000 pixels to drop any detection greater than the largest ground-truth object, and use morphological opening with structuring element of 5-pixel radius to eliminate small objects less than 10 pixels in width.

For our wide-area images, the four existing post-processing schemes do not sufficiently reduce detection errors. In this paper, we present a post-processing scheme consisting of an area thresholding sieving process followed by morphological closing. We use the VMO algorithm [3] (with the best recall rate among six detection algorithms we adapted in [7]) to test average F-score on the proposed scheme in contrast to the other four post-processing schemes, then present the performance evaluations on each of the five algorithms [2, 3, 4, 5, 6] in condition of before post-processing and using our post-processing scheme. Manual detection of vehicles of each frame in two datasets provides the ground truth objects. We classify the detection errors as false negatives and false positives, and quantify the detection performance of each algorithm without any post-processing and post-processed detections via precision, recall, F-score and percentage of wrong classification.

## 2. DATASETS

We use two datasets (720 × 480 pixels, 100 consecutive frames per dataset) for a performance comparison [7], [9] of each algorithm with and without a post-processing scheme. Both datasets represent wide-area aerial imagery with low spatial resolution. Traffic lanes were manually cropped from the aerial videos since the focus of this study is just the task on vehicle detection. (Road extraction can also be performed by GIS mapping or making use of an automatic detection method.) Rectangular shaped vehicles represent the ground truth in our manual segmentation. For the Tucson dataset, vehicle area is distributed from 40 to 150 pixels, and the total sum of vehicles is 4012 – i.e., an average of 40 vehicles per frame; for the Phoenix dataset, vehicle area ranges from 20 to 175 pixels, and the total sum of vehicles is 4060.

## 3. THE PROPOSED SCHEME

The proposed post-processing scheme has two stages: area-thresholding sieving and morphological closing. First, we sieve the detected objects by area thresholding (in pixels): a low threshold to exclude objects smaller than the smallest expected vehicle, and a high threshold that is larger than the largest expected vehicle size. All the binary objects within an area range, $A \in [t_{low}, t_{high}]$, are preserved. We used an area range of [5, 160] for the Tucson dataset and [5, 180] for the Phoenix dataset (not the same due to a slight difference in spatial resolution).

Since some detection errors may persist even after applying area thresholding, we include a second stage of post-processing – i.e., a morphological closing operation for connecting adjacent small objects that tend to be false detections, smoothing the border of each detection, and filling the tiny holes inside each detection.

## 4. PERFORMANCE METRICS

To evaluate the performance [13] of each of the object detection algorithms, we perform 8-connected component labeling of the binary detection output, then compare the overlap between detected objects and ground truth objects. Each detection can be classified as follows [7]:

True positive (TP): correct detection. If multiple detections intersect the same ground truth object, then only one TP is counted (the one having largest overlap). If a single detection intersects multiple ground truth objects, then only one TP is counted (the one having largest overlap);

False negative (FN): detection failure, indicated by a ground truth object failing to intersect any detected object;

False positive (FP): incorrect detection, indicated by a detection that does not intersect any ground truth object;

We use several evaluation metrics [7], [14] to quantify the detection performance:

Precision: a correctness measure, which is the ratio of TP to the sum of TP and FP;

Recall: a completeness measure, which is the ratio of TP to the sum of TP and FN;

F-score: the harmonic mean of precision and recall;

Percentage of wrong classification (PWC): the ratio of the sum of FP and FN to the sum of TP, FP, FN and TN; since we have no negative samples for detection, TN = 0.

## 5. EXPERIMENTS

We evaluated the quantitative detection performance in two scenarios: (1) the VMO algorithm combined with each of the five post-processing schemes, (2) five object detection algorithms combined with the proposed post-processing scheme. We also present a visual comparison of the detection performance.

### 5.1. VMO combined with five post-processing schemes

In the first experiment, we randomly picked up ten frames from each of the two datasets for performance evaluation. We combined the VMO algorithm with each of the five post-processing schemes, and for each post-processing scheme we selected a threshold scaling factor which was used in VMO to achieve the highest average F-score. The number of iterations in VMO was set to 10 for the Tucson dataset and 20 for the Phoenix dataset. Table I displays a comparison of average F-scores of VMO detection for the five schemes. The proposed scheme achieves the highest average F-score with tight confidence interval for both datasets.

TABLE I
AVERAGE F-SCORE OF VMO FOR FIVE POST-PROCESSING SCHEMES
(MEAN AND 95% CONFIDENCE INTERVAL)

| Post-Processing | Tucson Dataset | Phoenix Dataset |
|---|---|---|
| No Post-Processing | 0.504 ± 0.052 | 0.455 ± 0.030 |
| Filtered dilation | 0.781 ± 0.035 | 0.722 ± 0.034 |
| Heuristic Filtering | 0.744 ± 0.041 | 0.639 ± 0.041 |
| Filtering by SI | 0.557 ± 0.042 | 0.501 ± 0.024 |
| Sieving and Opening | 0.581 ± 0.067 | 0.674 ± 0.040 |
| Proposed Scheme | 0.787 ± 0.032 | 0.737 ± 0.028 |

We observed some shortcomings of the four existing post-processing schemes (with optimized thresholds for each scheme) for our wide-area images. Although the Filtered dilation scheme [11] improves the detection performance, the dilation operation expands the size of detected objects. Heuristic filtering [8] is efficient for post-processing only if vehicles have an aspect ratio very distinct from that of the false positives. The SI filtering scheme [12] results in a high number of remaining false detections. Zheng et al. [9] explained that a weakness of their sieving strategy is that objects can be wrongly excluded if they are too close to other objects having similar intensity (e.g., other vehicles, medians dividing lanes of traffic, or shadows). When applying Zheng's post-processing scheme [9] to our wide-area images, we find that morphological opening only works with a structuring element with 1- or 2-pixel radius to preserve correct detections since the vehicle width is only a few pixels.

We also investigated the sensitivity of the tuning parameters for the VMO algorithm. Fig. 1 shows the average F-score vs. the number of iterations used in VMO, and the best overall results are achieved using 10 iterations for the Tucson dataset and 20 iterations for the Phoenix dataset. Fig. 2 shows the average F-score vs. the threshold scaling factor used in VMO, and the best overall results are achieved via multiplying the optimum threshold surface [3] by a threshold scaling factor in the range of 0.6 to 0.7.

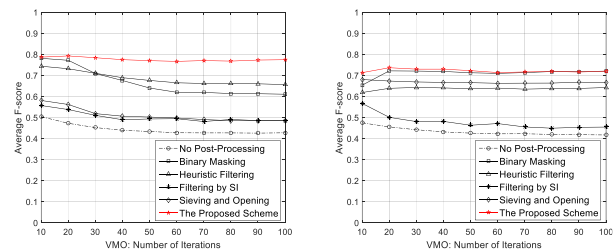

Fig. 1. Average F-score of VMO vs. number of iterations (Left: Tucson dataset; Right: Phoenix dataset).

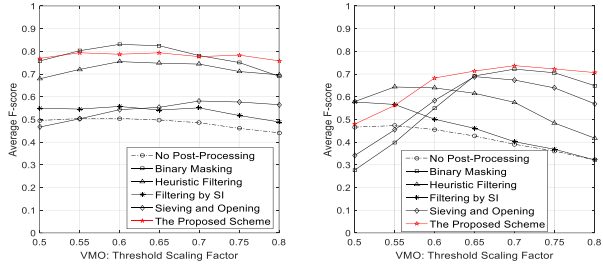

Fig. 2. Average F-score of VMO vs. threshold scaling factor (Left: Tucson dataset; Right: Phoenix dataset).

### 5.2. Five detection algorithms combined with our scheme

In the second experiment, we evaluated five object detection algorithms combined with the proposed post-processing scheme. We used a $3 \times 3$ structuring element to perform the closing operation for all of the results presented in this experiment. The detection algorithms that we were inspired by previously published works are the spectrum residual (SR) model [2], VMO algorithm [3], the fuzzy logic (FL) vehicle detection [4], feature density estimation (FDE) [5], and subpixel location with partial area effects (SL-PAE) [6]. The detection performance was evaluated using all 100 frames from each dataset. Table II shows the number of true positives (TP), false negatives (FN), and false positives (FP) before and after applying the proposed post-processing scheme. The results confirm that the proposed post-processing scheme significantly reduces the FP count for all five detection algorithms – as much as a 94% reduction in the case of FDE for the Tucson dataset. In exchange for decreasing the FN count, there is usually a FN increase and a TP decrease. The average FN count remains less than 5 per frame in the Tucson dataset and less than 10 per frame in the Phoenix dataset. The TP difference varies from a low decrease of 2.56% (FL, Phoenix dataset) to a high decrease of 20.77% (SL-PAE, Phoenix dataset) in addition to a 6.14% TP increase of FL algorithm in the Tucson dataset.

TABLE II

CLASSIFICATION OF DETECTIONS BEFORE AND AFTER THE PROPOSED POST-PROCESSING FOR TUCSON (T) AND PHOENIX (P) DATASETS

| Detection | | Algorithm | SR | VMO | FL | FDE | SL-PAE |
|---|---|---|---|---|---|---|---|
| TP | Before | T | 3584 | 3876 | 3207 | 3515 | 3739 |
| | | P | 2674 | 3420 | 3045 | 2965 | 3298 |
| | After | T | 3144 | 3126 | 3404 | 3007 | 3205 |
| | | P | 2259 | 3021 | 2967 | 2501 | 2613 |
| FN | Before | T | 180 | 49 | 175 | 418 | 258 |
| | | P | 1022 | 519 | 959 | 815 | 626 |
| | After | T | 537 | 82 | 231 | 648 | 560 |
| | | P | 1269 | 410 | 978 | 950 | 968 |
| FP | Before | T | 4189 | 7654 | 3244 | 3201 | 9100 |
| | | P | 4933 | 10836 | 18649 | 3505 | 7667 |
| | After | T | 686 | 3200 | 2254 | 183 | 1621 |
| | | P | 1947 | 2124 | 6900 | 1217 | 1603 |

Table III shows the percentage of wrong classification (PWC) for each detection algorithm combined with the proposed post-processing scheme. The least improvement was found with the FL algorithm, where the PWC reduced from 51.25% to 41.71% in the Tucson dataset, and reduced from 86.47% to 72.50% in the Phoenix dataset. The best performance is given by FDE, which has a PWC of 21.58% in the Tucson dataset. SR, FDE and SL-PAE exhibit better improvement for the Tucson dataset than the Phoenix dataset, while VMO behaves the opposite. The PWC for FDE and SL-PAE reduced to below 50% for both datasets.

TABLE III

PERCENTAGE WRONG CLASSIFICATION BEFORE AND AFTER PROPOSED POST-PROCESSING
(MEAN AND 95% CONFIDENCE INTERVAL)

| PWC % | | Algorithm | SR | VMO | FL | FDE | SL-PAE |
|---|---|---|---|---|---|---|---|
| Before | T | | 54.39 ± 1.08 | 65.60 ± 1.09 | 51.25 ± 0.96 | 50.41 ± 0.88 | 71.45 ± 0.62 |
| | P | | 68.35 ± 0.95 | 76.56 ± 0.55 | 86.47 ± 0.28 | 59.30 ± 0.99 | 71.34 ± 0.59 |
| After | T | | 27.78 ± 1.16 | 50.83 ± 0.98 | 41.71 ± 1.27 | 21.58 ± 0.77 | 40.25 ± 0.90 |
| | P | | 58.42 ± 0.89 | 45.18 ± 1.07 | 72.50 ± 0.58 | 46.07 ± 1.29 | 49.26 ± 1.19 |

Fig. 3 shows precision, recall and F-score for each algorithm before (left) and after (right) the proposed post-processing. Before post-processing, VMO exhibits the highest recall for both datasets, the second lowest precision and F-score (next to SL-PAE) for the Tucson dataset, and the second lowest precision and F-score (next to FL) for the Phoenix dataset. FDE has the highest precision and F-score for both datasets, the lowest recall for the Tucson dataset, and the median recall for the Phoenix dataset.

After post-processing, each algorithm for both datasets achieves recall rates higher than 0.7 except for SR in the Phoenix dataset, demonstrating effectiveness of the proposed post-processing scheme. SR in the Tucson dataset and SL-PAE in both datasets display a decrease in recall value of only about 0.1 as a result of post-processing. For VMO, FL and FDE, the recall level is almost unaffected by the post-processing (within 0.08 in mean); VMO also preserves the best recall for both datasets. For the Tucson dataset, FDE has the highest precision and F-score, VMO has the lowest precision and F-score. For the Phoenix dataset, FDE has the best precision, VMO has the best F-score, and FL has the lowest precision and F-score. The statistical data is validated by the tight confidence intervals for each of the metrics.

### 5.3. Visual comparison of detection performance

We used a subimage with size $64 \times 64$ from the 50th frame of each dataset for a visual comparison of object detection performance. Fig. 4 shows the results of each detection algorithm before and after post-processing, where detected objects are bounded in color, and the bottom row shows ground-truth (GT) labeling. Comparison of the result columns before post-processing and after post-processing illustrates the efficient removal of FPs by the proposed post-processing scheme.

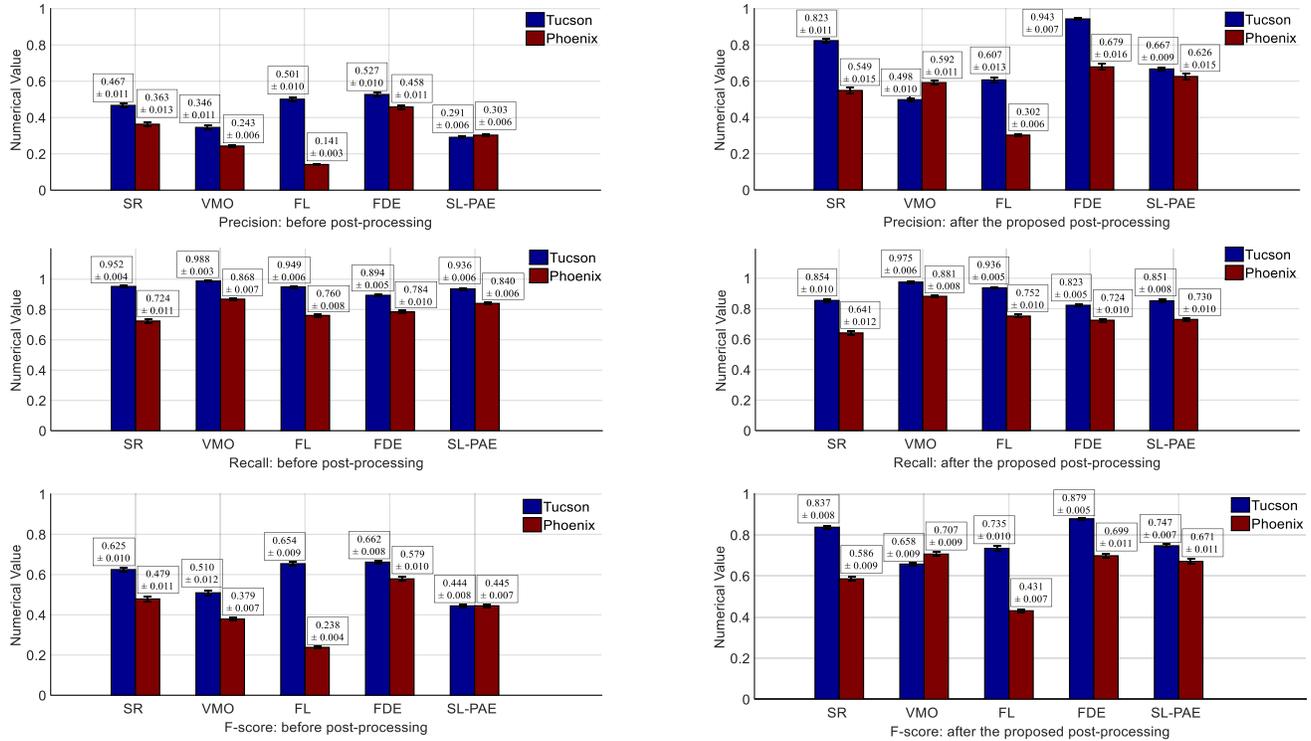

Fig. 3. Comparison of precision, recall and F-score for each algorithm: before (left) and after (right) the proposed post-processing.

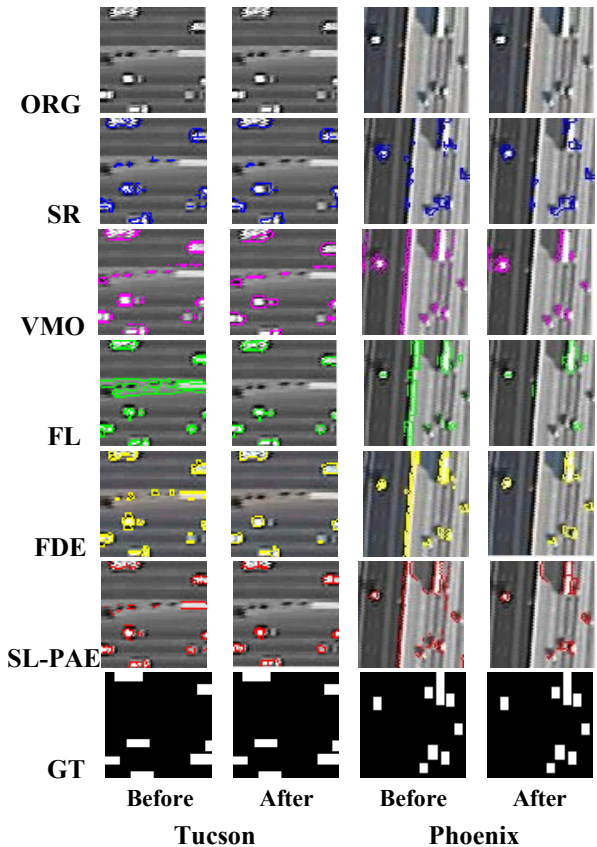

Fig. 4. Original images (ORG) and visual comparison of vehicle detections via the five object detection algorithms before and after the proposed post-processing scheme.

## 6. CONCLUSION

We have presented a two-stage post-processing scheme comprising area-thresholding object sieving followed by morphological closing. Experimental results on two image datasets demonstrate that this scheme outperforms four other post-processing schemes through average F-score comparison. We use the proposed scheme to test five object detection algorithms [2, 3, 4, 5, 6] for two datasets, and the performance comparison demonstrates that the proposed scheme improves both precision and PWC, reduces FPs, and preserves recall for each algorithm. The proposed scheme on post-processing represents a useful tool for the performance enhancement of object detection methods in wide-area aerial imagery.

As future work, we plan to upgrade the proposed scheme by improving the performance of object sieving process with semi-soft area-thresholding in the presence of average vehicle size, which will reduce the loss of TPs that negatively affects recall; we also aim to present more sufficient proof for the merits of the proposed scheme via a performance analysis by a variety of selected schemes post-processed on the latest object detection algorithms.

## 7. ACKNOWLEDGMENT

The authors wish to thank Prof. Mark Hickman, School of Civil Engineering, The University of Queensland, Australia, for providing the two aerial video datasets.